# Markov model with machine learning integration for fraud detection in health insurance

Rohan Yashraj Gupta, Satya Sai Mudigonda, Pallav Kumar Baruah, Phani Krishna Kandala

*Abstract*: Fraud has led to a huge addition of expenses in health insurance sector in India. The work is aimed to provide methods applied to health insurance fraud detection. The work presents two approaches - a markov model and an improved markov model using gradient boosting method in health insurance claims. The dataset 382,587 claims of which 38,082 claims are fraudulent. The markov based model gave the accuracy of 94.07% with F1-score at 0.6683. However, the improved markov model performed much better in comparison with the accuracy of 97.10% and F1-score of 0.8546. It was observed that the improved markov model gave much lower false positives compared to markov model.

*Keywords: Health Insurance, Fraud detection, Markov model, Actuarial, Gradient boosting method.*

## I. INTRODUCTION

There has been a rapid growth in the insurance industry which led to an increase in the number of insurance claims considerably. According to a report by Insurance Business the Indian insurance market is growing at a rate of 14.5% [1]. An insurance company, by its nature, is very susceptible to fraud. One such industry is health insurance.

Fraud is one of the major problem in the health insurance industry which causes significant losses. Melih et.al define insurance fraud as "knowingly making a fictitious claim, inflating a claim or adding extra items to a claim, or being in any way dishonest with the intention of gaining more than legitimate entitlement" [2]. According to the 2019 report of National Health Care Anti-Fraud Association on healthcare fraud detection, the total losses in 2018 was USD 679.18 million which is expected to reach USD 2.54 billion by 2024 (Health insurance fraud and its impact on the healthcare system).

Traditionally, fraud detection was done with rule-based systems which required great involvement of domain experts [3]. In recent times, we can find many works that use data science techniques such as data mining [3], [4], machine learning [5][6]–[9], social network analysis [10]–[12], etc to a great extent.

In this work, we proposes a model for fraud detection that uses concepts of markov model. The model is tested on health insurance dataset. The proposed model shows a significant improvement when a machine learning model is incorporated into it.

The work is divided into 6 sections, section 1 gives the introduction. Section 2 presents the literature available currently in similar work. Section 3 details the background of the markov model. Section 4 explains the experimental setup of this work. Section 5 presents the results and discussion. Section 6 provides the conclusion and the future work that is being done in this area.

## II. LITERATURE REVIEW

Fraud, anomaly or intrusion detection are terms used to define the problem of finding unusual patterns or activities in the data. Researchers are constantly finding better ways to tackle this problem and in this pursuit many methods have been developed. Various models have been developed to tackle this problem in numerous domains such as server systems [13], network [14], electronic systems [15], Insurance [16]–[18], banking sector [8], etc.

Hossein Joudaki et.al (2015) has used various data mining methods to find fraud and abuse in healthcare sector. They recommend seven steps methodology for finding fraud in healthcare claims [19]. Qi Liu et.al (2013) have worked on health care data and used datamining and machine learning techniques to analyse the data. The work majorly focuses on Medicaid and Medicare sector. They have used clustering method to distinguish the claim distributions arising from different diseases which they use to find fraudulent claims [20]. Rohan et.al [16] have proposed a framework for fraud detection, which incorporates various actuarial and data science techniques. They have also implemented gradient boosting method for fraud detection in motor insurance dataset [21]. Nikhil et.al have built a machine learning based on a globally available motor insurance dataset[22]. A group of researcher from the Society of Actuaries have performed a comprehensive study in healthcare fraud. They have analysed 470 papers in healthcare sector and identified 27 most relevant research papers and articles in this domain [23].

Many work can be seen where markov models have been used in the areas like anomaly detection, outlier detection, fraud detection, etc. Evaristo et. al. have developed an actuarial statistical model using markov models which is applied in health care sector. They have used this to predict the future cost [18]. Sultana et. al. have used hidden markov

**Authors' Detail**
\* Correspondence Author

  **Rohan Yashraj Gupta\*,** Department of Mathematics and Computer Science, Sri Sathya Sai Institute of Higher Learning, Puttaparthi, India. Email: rohanyashrajgupta@sssihl.edu.in

  **Satya Sai Mudigonda,** Department of Mathematics and Computer Science, Sri Sathya Sai Institute of Higher Learning, Puttaparthi, India. Email: satyasaibabamudigonda@sssihl.edu.in

  **Pallav Kumar Baruah,** Department of Mathematics and Computer Science, Sri Sathya Sai Institute of Higher Learning, Puttaparthi, India. Email: pkbaruah@sssihl.edu.in

  **Phani Krishna Kandala,** Department of Mathematics and Computer Science, Sri Sathya Sai Institute of Higher Learning, Puttaparthi, India. Email: kandala.phanikrishna@gmail.com

model for anomaly detection. Host-based anomaly detection technique is used as an approach which ensures the safety and security of systems. The work contributes a method which has a significantly lesser training time of the model [24]. Vasheghani et. al. have used time series anomaly detection method using markov chains. The model is tested on medical data, utility usage data and New York tax data [25].

The motivation for this work is derived from the fact in literature there is no work where markov models are used in insurance fraud detection.

## III. MARKOV MODEL BACKGROUND

If the probabilities for the future values of a process are dependent only on the latest available value, the process has the Markov property. Mathematically, for a process with time set {1, 2, 3, ...} and a discrete state space :

$$P(X_n = x_n \mid X_{n-1} = x_{n-1}, X_{n-2} = x_{n-2}, \ldots, X_1 = x_1)$$
$$= P(X_n = x_n \mid X_{n-1} = x_{n-1})$$

For a continuous-time process with a discrete state space, we need to express this in the form:

$$P(X_n = x_n \mid F_s) = P(X_n = x_n \mid X_s)$$

For a continuous-time process with a continuous state space, we need to express this in the form:

$$P(X_n \in A \mid F_s) = P(X_n \in A \mid X_s)$$

## IV. EXPERIMENTAL SETUP

### A. Data Description

The data used for this work is a health insurance dataset. This consists of claims and policy data for one policy year starting 20th August, 2019. The table below summarizes the dataset.

| Status | Count | % of total |
|---|---|---|
| Fraud | 38,082 | 9.95% |
| Not-Fraud | 344,505 | 90.05% |
| Grand Total | 382,587 | 100.00% |

The dataset has a total of 26 features. The description of the data is given in the table below.

| Feature | Description |
|---|---|
| Policy Number | Unique policy identification number |
| Insured Id | Unique ID given to insured |
| Claim Identification Number | Unique claim identification number |
| Benefit Type | Type of benefit (medical / surgical) |
| Claim Status | Current status of the claim |
| Treatment Start Date | Treatment Start Date |
| Treatment End Date | Treatment End Date |
| Claim Settlement Date | Date at which the claims were settled |
| Claim Reported Date | Date at which the claims were reported |
| Claim Billed Amount | Claims amount billed |
| Approved / Allowed Amount | Approved / Allowed Amount |
| Claim Paid Amount | Claims amount paid |
| Medical Service Provider ID | Unique ID given to hospitals |
| Medical Service Provider Name | Name of the hospital |
| No of Days Stayed | Days stayed in the hospital |
| Primary Diagnosis Code | Unique code given to diagnosis |
| Primary Diagnosis Name | Name of the diagnosis |
| Primary Procedure Code | Unique code given to procedure |
| Primary Procedure Name | Name of the procedure |
| Net Amt | Net amount paid to the insured |
| Claim Paid Date | Date of payment of the claim |
| Surgery Date | Date of surgery |
| Discharge Date | Date of discharge |
| Claim Raised Date | Date when the claim was raised |
| Hospital District | District where the hospital is located |
| Claim Status | Status of the claims as fraud / not-fraud |

### B. Markov Model

Each of the features in the dataset used for this purpose was categorized to into groups based on quantiles, such that the groups had equal number of claims in it [26], [27], [24]. The sequence of the values taken in the feature was labelled into states. For e.g. for a dataset with three features where each of the feature can take three values the total number of possible states would be 27. Similarly, in this work we have considered five most significant features and after doing the categorization of the features, the total number of states thus formed was 1,188. The snapshot of the labelled states are given in the table below:

| Benefit Type | No of Days Stayed | Primary Diagnosis Code | Hospital Type | Net Amt | States |
|---|---|---|---|---|---|
| MEDICAL | medium | M1 | Private | high | 1 |
| MEDICAL | medium | M1 | Private | medium | 2 |
| MEDICAL | short | M1 | Private | high | 3 |
| MEDICAL | medium | M1 | Public | medium | 4 |
| MEDICAL | long | M1 | Public | medium | 5 |
| MEDICAL | long | M1 | Private | high | 6 |
| MEDICAL | medium | M1 | Public | medium | 4 |
| MEDICAL | long | M3 | Private | high | 7 |
| MEDICAL | medium | M1 | Private | medium | 2 |
| MEDICAL | long | M1 | Private | high | 6 |
| … | … | … | … | … | … |
| SURGICAL | long | S5 | Private | high | 8 |

Each of the claim has a class label as fraud or not-fraud. Now, based on the data as shown in the table above a model was fit to determine the probability for a claim being fraudulent or not. Figure below shows the diagrammatic representation of the underlying model. In the figure, states 1 and 9 have been labelled with blue and orange respectively. Similarly, an extensive network can be visualized consisting of all the 1,188 states. All of 382,587 claims can be categorized into one of these states. Each of the states has the probability of it being fraudulent or not. Let us consider the probability of state 1 being fraudulent i.e. *P(Claim=Fraud | State = 1)*. So, the probability would be written as:

$P(Claim = Fraud | State = 1) =$
$P(Benefit\ Type = MEDICAL) * P(No\ of\ Days\ Stayed = medium | Benefit\ Type = medium) * P(Primary\ Diagnosis\ Code = M1 | Benefit\ Type = medium) * P(Hospital\ Type = Private | Primary\ Diagnosis\ Code = M1) * (Net\ Amt = high | Primary\ Diagnosis\ Code = M1) * (Claim\ Status = Fraud | Primary\ Diagnosis\ Code = M1)$

$P(Claim = Not\text{-}Fraud | State = 1) = 1 - P(Claim = Fraud | State = 1)$

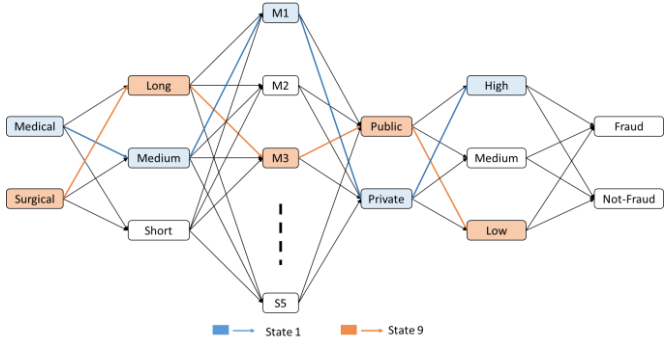

Similarly, we can find the respective probability for all the states. The model was built as explained. For this, the dataset was divided into train and test in the ratio of 70:30. The split in the dataset was made randomly and was used to determine the probabilities.

Using the method explained above the probability for each of the state being fraud was derived for the train dataset. The model was then tested on the test dataset. The confusion matrix below summarizes the performance of the model.

|  | Reference |  |
|---|---|---|
| **Prediction** | **Fraud** | **Not-Fraud** |
| **Fraud** | 6,857 | 2,120 |
| **Not-Fraud** | 4,687 | 1,01,113 |

Using the confusion matrix shown above, various performance metrics have been calculated as shown in the table below:

| Measure | Value | Derivations |
|---|---|---|
| Sensitivity | 0.5940 | TPR = TP / (TP + FN) |
| Specificity | 0.9795 | SPC = TN / (FP + TN) |
| Precision | 0.7638 | PPV = TP / (TP + FP) |
| Accuracy | 0.9407 | ACC = (TP + TN) / (P + N) |
| F1 Score | 0.6683 | F1 = 2TP / (2TP + FP + FN) |

We observe that the sensitivity of the model 0.59 meaning that 59% of the fraud cases were correctly identified. And a specificity of 0.97 means that 97% of the non-fraud cases were correctly identified. The accuracy of the model is 0.94, meaning that 94% of the labels were correctly identified by the model.

The figure below shows the Receiver Operating Characteristic (ROC) curve for the model. The ROC curve is the plot of the pair of sensitivity and specificity for different threshold values. The area under the curve (AUC) is the measure which is used as another useful metric which is the areas that the ROC curve cover. A good model should have the AUC closer to 1. The model proposed has the AUC of 0.8424.

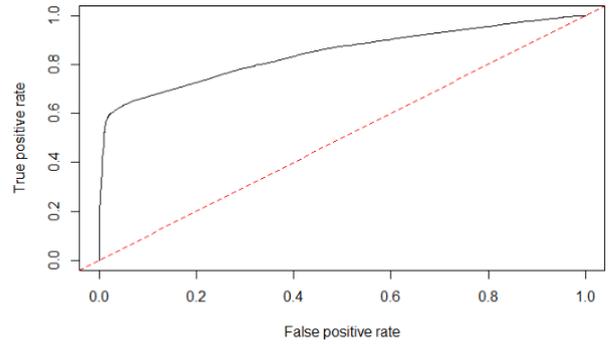

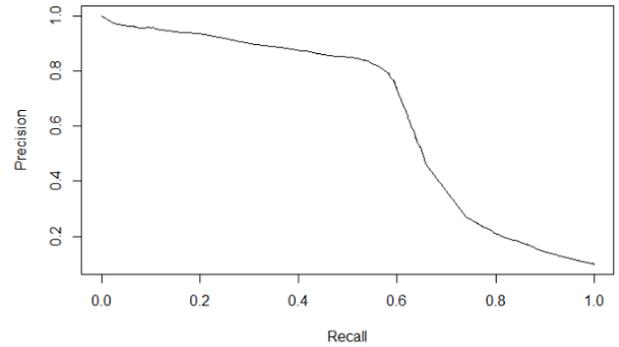

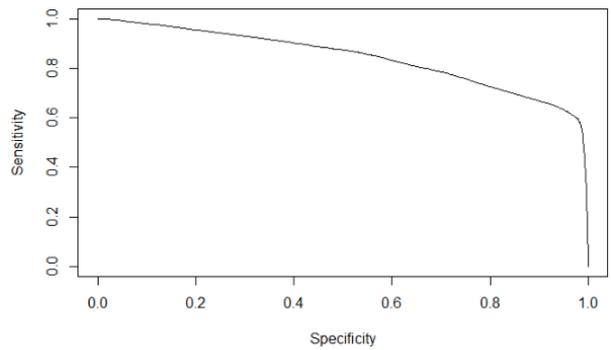

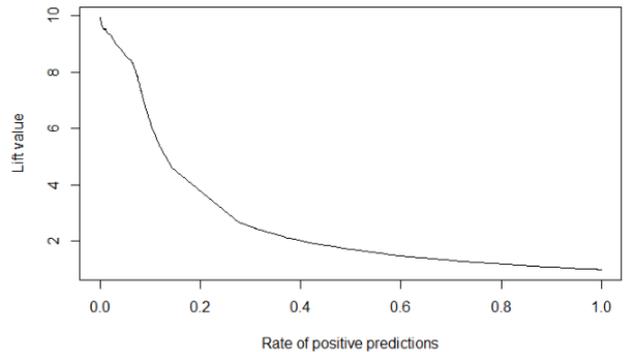

The model was further improved by including more features into the data and retaining the some of the features as it is i.e. no categorization was done. The details are given in the next section.

### C. Markov Model using GBM

One of the limitation while working with Markov Model was that the features had be categorized into buckets. This was required otherwise the total number of states would be too high for the markov model. However, with the use of machine learning model, we can work with a very high number of states and the associated probabilities can be learnt for all the states [21].

In this model, three additional features are taken into consideration, viz. "Medical Service Provider ID", "Hospital District" and "Amount paid to Hospital". The data is divided into train and test in the ratio 70:30. Gradient Boosting Method is used for building a fraud detection model.

For the GBM modelling a total of 300 trees were used. The maximum depth of each tree (i.e., the highest level of variable interactions allowed) was kept as 5. Learning rate or step-size reduction was kept at 0.1. In addition to the usual fit, a 10-fold cross-validation was performed.

From the figure, it can be seen how the improvement in the model took place with every iteration. The dotted blue line shows the point where the Bernoulli deviance of the model converged and there was no further significant reduction in the model.

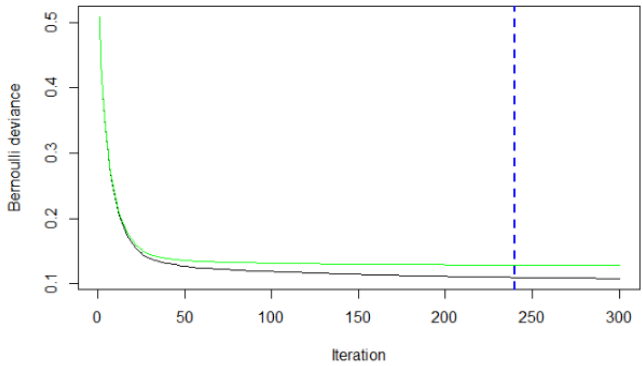

The model was tested on test data. The confusion matrix and statistics below summarizes the performance of the model.

|  | Reference | |
|---|---|---|
| **Prediction** | **Fraud** | **Not-Fraud** |
| **Fraud** | 9,796 | 1,586 |
| **Not-Fraud** | 1,748 | 1,01,647 |

The confusion matrix was used to calculate various metrics like sensitivity, specificity, precision, accuracy, F1-score. This is shown in the table below:

| **Measure** | **Value** |
|---|---|
| Sensitivity | 0.8486 |
| Specificity | 0.9846 |
| Precision | 0.8607 |
| Accuracy | 0.9710 |
| F1 Score | 0.8546 |

It can be observed that the sensitivity is now at 0.84
Area under the curve: 0.9926

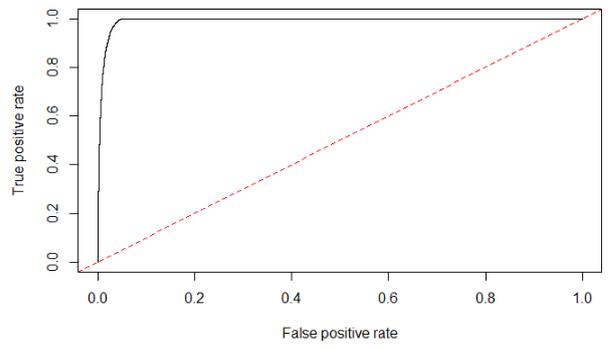

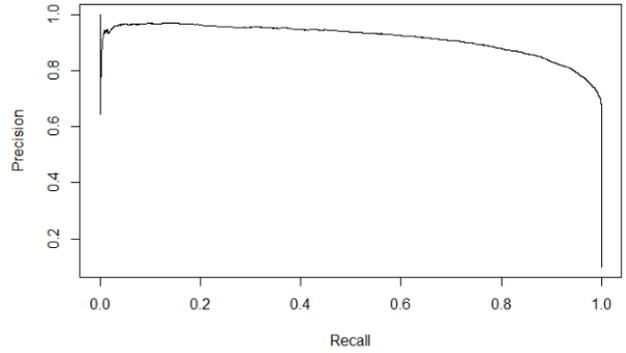

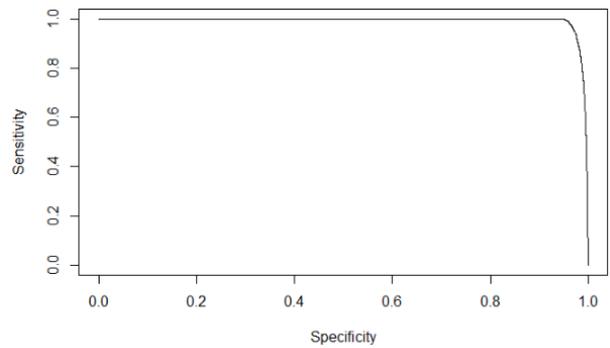

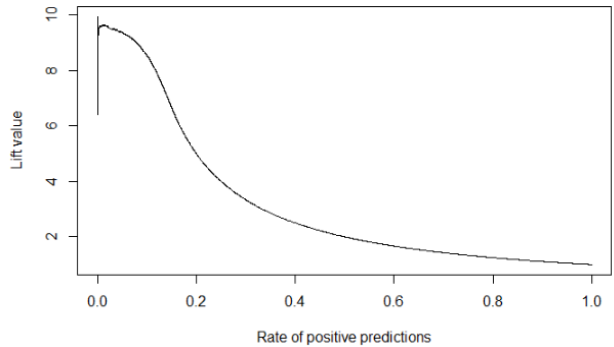

## V. RESULTS AND DISCUSSION

From the results obtained from both the methods implemented, as discussed in the previous sections, it can be observed that the idea of markov model can be extended using a machine learning model. In this work, we have improved the markov model using GBM. The performance of improved morkov model was significantly better than the markov model.

## VI. CONCLUSION AND FUTURE WORK

In this work, we have built a fraud detection model using markov model. The model was tested on a health insurance dataset. One major drawback of this model was that the number of states under consideration was too less for this purpose. The sensitivity of the model was too less at 0.59, specificity was 0.98 and the F1-score of the model was 0.66.

The model was improved further by the use of Gradient Boosting Method. For this purpose, more features were added into the dataset, thus the total number of states now formed were much higher than the previous cases. Under this scenario when the model was re-run, there was a significant improvement in the model. The sensitivity of the model was now 0.84, specificity was 0.98 and the F1-score was 0.85.

We can thus conclude that when a machine learning model is incorporated into some of the statistical models (in this case markov model), we can expect a significant improvement in the performance.

The work has been performed in health insurance business. However, this can be extended into other lines of businesses too like life insurance, motor insurance, etc. In the future we will see how the model performs for other lines of business.

## VII. Acknowledgement